\documentclass[runningheads]{llncs}
\usepackage{graphicx}
\usepackage[normalem]{ulem} 
\usepackage{amsfonts}
\usepackage{caption}
\usepackage{subcaption}
\begin{document}

\title{GAN Path Finder: Preliminary results}

\author{Natalia Soboleva\inst{1, 2} \orcidID{0000-0002-9071-7456} \and
Konstantin Yakovlev\inst{1, 2}\orcidID{0000-0002-4377-321X}}

\authorrunning{N. Soboleva and K. Yakovlev}

\institute{National Research University Higher School of Economics, Moscow, Russia\\
\email{nsoboleva@edu.hse.ru, kyakovlev@hse.ru}\\
\and
Federal Research Center ``Computer Science and Control'' of Russian Academy of Sciences, Moscow, Russia\\
\email{yakovlev@isa.ru}}
\maketitle 
\begin{abstract}
2D path planning in static environment is a well-known problem and one of the common ways to solve it is to 1) represent the environment as a grid and 2) perform a heuristic search for a path on it. At the same time 2D grid resembles much a digital image, thus an appealing idea comes to being -- to treat the problem as an image generation task and to solve it utilizing the recent advances in deep learning. In this work we make an attempt to apply a generative neural network as a path finder and report preliminary results, convincing enough to claim that this direction of research is worth further exploration.

\end{abstract}

{\let\thefootnote\relax\footnote{{Camera-ready version of the paper as to appear in KI'19 proceedings}}}

\section{Introduction}
Grids composed of blocked and free cells are commonly used to represent static environment of a mobile agent. They appear naturally in game development \cite{sturtevant2012benchmarks} and are widely used in robotics \cite{elfes1989using}, \cite{thrun2003learning}. When the environment is represented by a grid, heuristic search algorithms, e.g. A* \cite{hart1968formal}, are typically used for path planning. These algorithms iteratively explore the search space guided by a heuristic function such as Euclidean or octile distance. When the obstacles are present on the way such guidance leads to unnecessary exploration of the areas surrounding the obstacles. This issue can be mitigated to a certain extent by weighting the heuristics \cite{Ebendt:2009:WAS:1576851.1576888}, using random jumps \cite{rstar} or skipping portions of the search space exploiting the grid-induced symmetries \cite{Harabor2011OnlineGP}. At the same time, having in mind, that grids resemble digital images a lot and recently convolutional neural networks demonstrate tremendous success various image processing tasks, an orthogonal idea can be proposed -- to plan entirely in the image domain using the state-of-the-art deep learning techniques thus avoiding the unnecessary state-space exploration by construction. In this work we leverage this idea and report preliminary results on path finding as image generation. We describe generative adversarial net that generates a path image in response to context input, i.e. image of the grid-map with start and goal. We demonstrate empirically that the proposed model can successfully handle previously unseen instances.

\section{Related work}

\begin{figure}[t]
\centering
\begin{subfigure}[b]{0.24\linewidth}
    \centering
    \includegraphics[width=\linewidth]{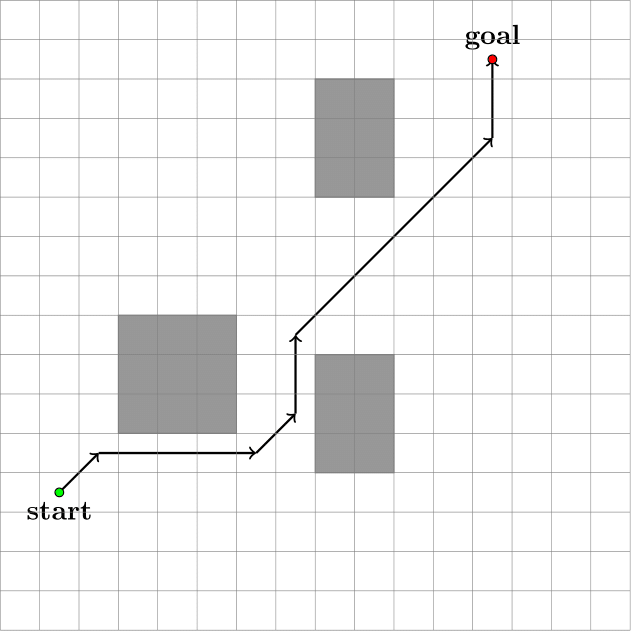}
    \caption{grid}
\end{subfigure}
\begin{subfigure}[b]{0.24\linewidth}
    \centering
    \includegraphics[width=\linewidth]{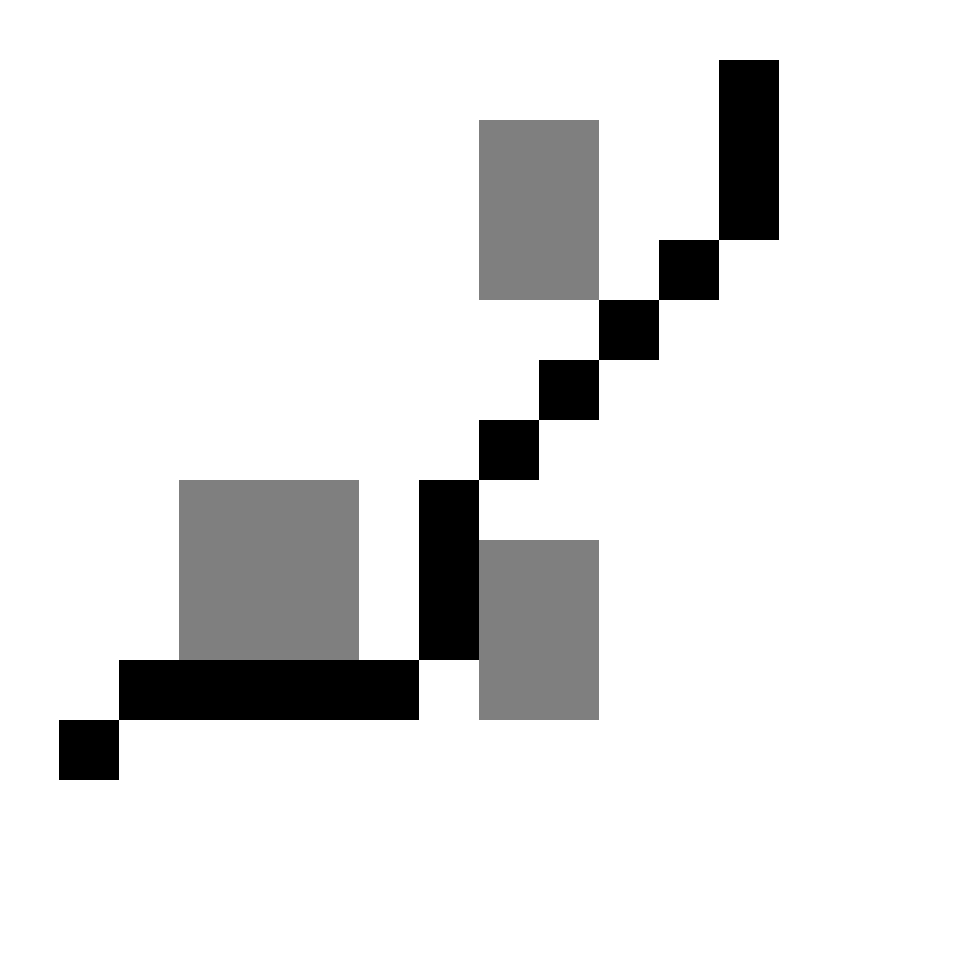}
    \caption{ground truth}
\end{subfigure}
\begin{subfigure}[b]{0.24\linewidth}
    \centering
    \includegraphics[width=\linewidth]{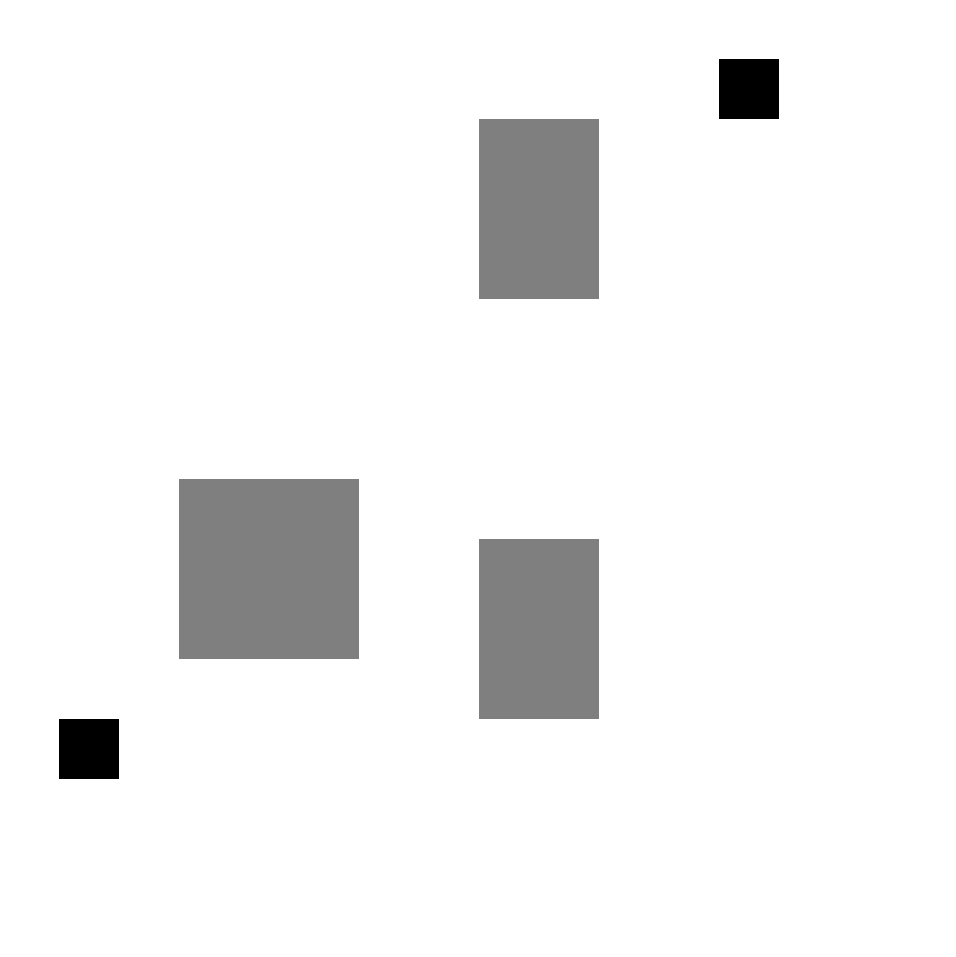}
    \caption{input}
\end{subfigure}
\begin{subfigure}[b]{0.24\linewidth}
    \centering
    \includegraphics[width=\linewidth]{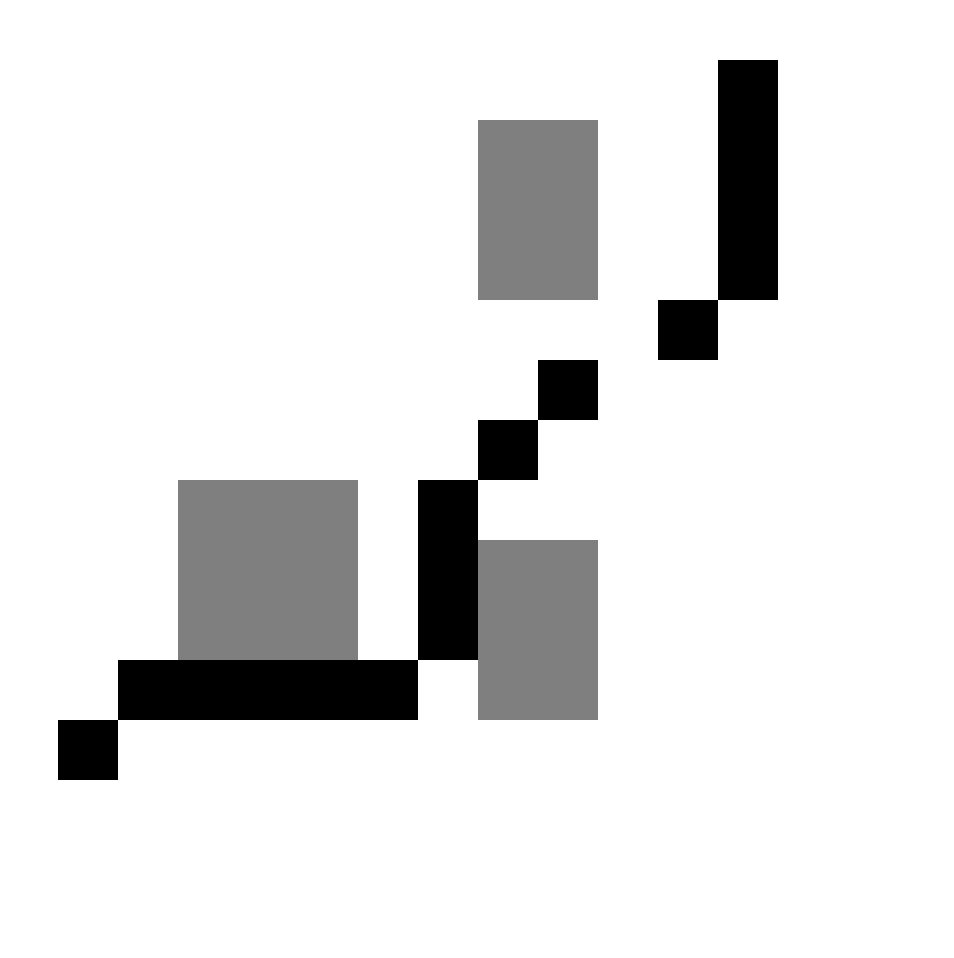}
    \caption{output}
\end{subfigure}
\caption{a) A $16 \times 16$ grid and a path on it; b) corresponding image; c) image-input for the generator; d) image-ouput of the generator. For b), c), d) image pixels are depicted as squares for illustrative purposes.}
\label{fig_problem_statement}
\end{figure}

The line of research that is most relevant to our work is deep learning (DL) for path/motion planning. A wide variety of works, e.g. \cite{Finn2017motionplanning}, \cite{eitel2017learning}, are focused on motion planning for manipulators. Unlike these works we are interested in path planning for mobile agents. DL approaches to navigation in 3D environment, that rely on the first-person imagery input, are considered in \cite{Gupta_2017_CVPR}, \cite{zhu2017target}, \cite{mirowski2018learning}. In contrast to these papers we focus on 2D path finding when a top-down view, i.e. a grid-map, is given as the input. In \cite{tamar2016value} such a task was considered, among the others, when the Value Iterations Networks (VINs) were presented. Evaluation was carried out on $16 \times 16$ and $28 \times 28$ grids. We are targeting larger maps, i.e. $64 \times 64$ grids. In \cite{lee2018gated} it was shown that VINs are ``often  plagued  by training instability, oscillating between high and low performance between epochs'' and other drawbacks. Instead Gated Path Planning Networks (GPPNs) were proposed but, again, the evaluation was carried out only on grids of size $15 \times 15$ and $28 \times 28$. The most recent work on VINs \cite{vins_new128} proposes a pathway to use them on larger maps via abstraction mechanism,value iteration network is applied to $8 \times 8$ feature maps. Unlike VINs or other approaches based on reinforcement learning (e.g. \cite{panov2018grid}), this work \textit{i}) is not rooted in modeling path planning with Markov decision process, \textit{ii}) considers quite large grids, i.e. $64 \times 64$, as the indecomposable input. 

\section{Problem statement}

Consider a 2D grid composed of blocked and unblocked cells with two distinguished cells -- start and goal. The path on a grid is a sequence of adjacent unblocked cells connecting start and goal\footnote{we are assuming 8-connected grids in this work.} -- see \figurename \ref{fig_problem_statement}a. The task of the path planner is to find such path. Often the length of the path is the cost objective to be minimized, but in this work we do not aim at finding shortest paths.

Commonly the task is solved by converting the grid to the undirected graph and searching for a path on this graph. Instead we would like to represent the grid as an image and given that image, generate a new one that implicitly depicts the path -- see \figurename \ref{fig_problem_statement}.

\section{GAN Path Finder}

\subsubsection{Grid-to-image conversion} To convert the grid to an image we use a straightforward approach: we distinguish 3 classes of the cells, -- free, blocked and path (incl. start and goal), and assign a unique color to each of them. Although, we do not use loss functions based on the pixel-distance later on, we intuitively prefer this distance to be maximal between the free pixels and the path pixels. Thus, the free cells become the white pixels of the grayscale image, path cells (including start and goal) -- black, blocked cells -- gray (as depicted on \figurename \ref{fig_problem_statement}b,c,d).

\subsubsection{Types of images} 3 types of images depicting grids and paths are to be distinguished. First, the image that depicts the grid with only start and goal locations -- see \figurename \ref{fig_problem_statement}c. This is the input. Second, the generated image that is the output of the neural network we are about to construct -- see \figurename \ref{fig_problem_statement}d. Third, the ground truth image that is constructed by rendering the input image with the A* path on it -- see \figurename \ref{fig_problem_statement}a. Ground truth images are extensively used for training, i.e. they serve as the examples of how ``good'' images look like. At the same time, from the path-finding perspective we should not penalize the output images that differ from the corresponding ground-truth image but still depict the correct path from start to goal. We will cover this aspect later on.

\subsubsection{Architectural choices} Convolutional neural networks (CNNs) \cite{Krizhevsky2012ImageNetCW} are the natural choice when it comes to image processing tasks. As there may exist a few feasible paths on a given grid we prefer to use Generative Adversarial Nets (GANs) \cite{2014GANs} for path planning as we want our network to learn some general notion of the feasible path rather than forcing it to construct the path exactly in the same way that the supervisor (e.g. A*) does. GAN is composed of 2 sub-networks -- generator and discriminator. Generator tries to generate the path-image, while discriminator is the classifier that tries to tell whether the generated image is ``fake'', i.e. does not come from the distribution of ground-truth images. Both these networks are the CNNs that are trained simultaneously.

In this work we are, obviously, not interested in generating the images that depict some random grid with a correct path on it, but rather the image that depicts the solution of the given path finding instance (encoded as the input image). That leads us to the so-called conditional GANs (cGANs) \cite{DBLP:journals/corr/MirzaO14}. Conditioning here means that the output of the network should be conditioned on the input (and not just generated out of the random noise as in vanilla GANs). We experimented with two prominent cGAN architectures -- Context Encoders \cite{2016CE} and pix2pix \cite{DBLP:journals/corr/IsolaZZE16}. CE -- is the model that is tailored to image inpainting, i.e. filling the missed region of the image with the pixels that look correct. In our case we considered all the free-space pixels as missing and make CE inpaint them. Pix2pix is a more general cGAN that is not tailored to particular generation task but rather solves the general ``from pixels to pixels'' problem. In path finding we want some free pixels become path pixels. We experimented with both CE and pix2pix and the latter showed more convincing results, which is quite foreseeable as pix2pix is a more complex model utilizing residual blocks. Thus, we chose pix2pix as the starting point for our model. 

\begin{figure}[t]
\centering
\begin{subfigure}{0.24\textwidth}
    \centering
    \includegraphics[width=\linewidth]{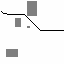}
    \caption{ground truth}
\end{subfigure}
\begin{subfigure}{0.24\textwidth}
    \centering
    \includegraphics[width=\linewidth]{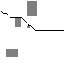}
     \caption{generated}
\end{subfigure}
\begin{subfigure}{0.24\textwidth}
    \centering
    \includegraphics[width=\linewidth]{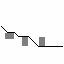}
    \caption{ground truth}
\end{subfigure}
\begin{subfigure}{0.24\textwidth}
    \centering
    \includegraphics[width=\linewidth]{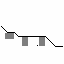}
     \caption{generated}
\end{subfigure}
\caption{Examples of the generated solutions, for which the MSE metric is not consistent, as the generated paths do not match ground truth, but are still feasible.}
\label{fig_diff_paths}
\end{figure}

\subsubsection{Generator}

The general structure of generator is borrowed from pix2pix \cite{DBLP:journals/corr/IsolaZZE16}. The difference is that in original work authors suggested two slightly different types of generator: the one with separated encoder/decoder parts with residual blocks in the bottleneck and the one utilizing skip-connections through all layers following the general shape of a ``U-Net'' architecture \cite{DBLP:journals/corr/RonnebergerFB15}. We experimented with both variants and the latter appeared to be more suitable for the considered task.

Original pix2pix generator's loss function is the weighted sum of two components. The first component penalizes generator based on how close the generated image is to the ground truth one, the second one -- based on the discriminator's response (adversarial loss). We kept the second component unchanged while modified the first one to be the cross-entropy rather then the L1 pixel-distance. The rationale behind this is that in the considered case we prefer not to generate a color for each pixel but rather to classify whether it belongs to ``free'', ``blocked'', ``path'' class. This turned to provide a substantial gain in the solution quality.

\begin{figure}[t]
\centering
\begin{subfigure}{0.3\linewidth}
    \centering
    \includegraphics[width=\linewidth]{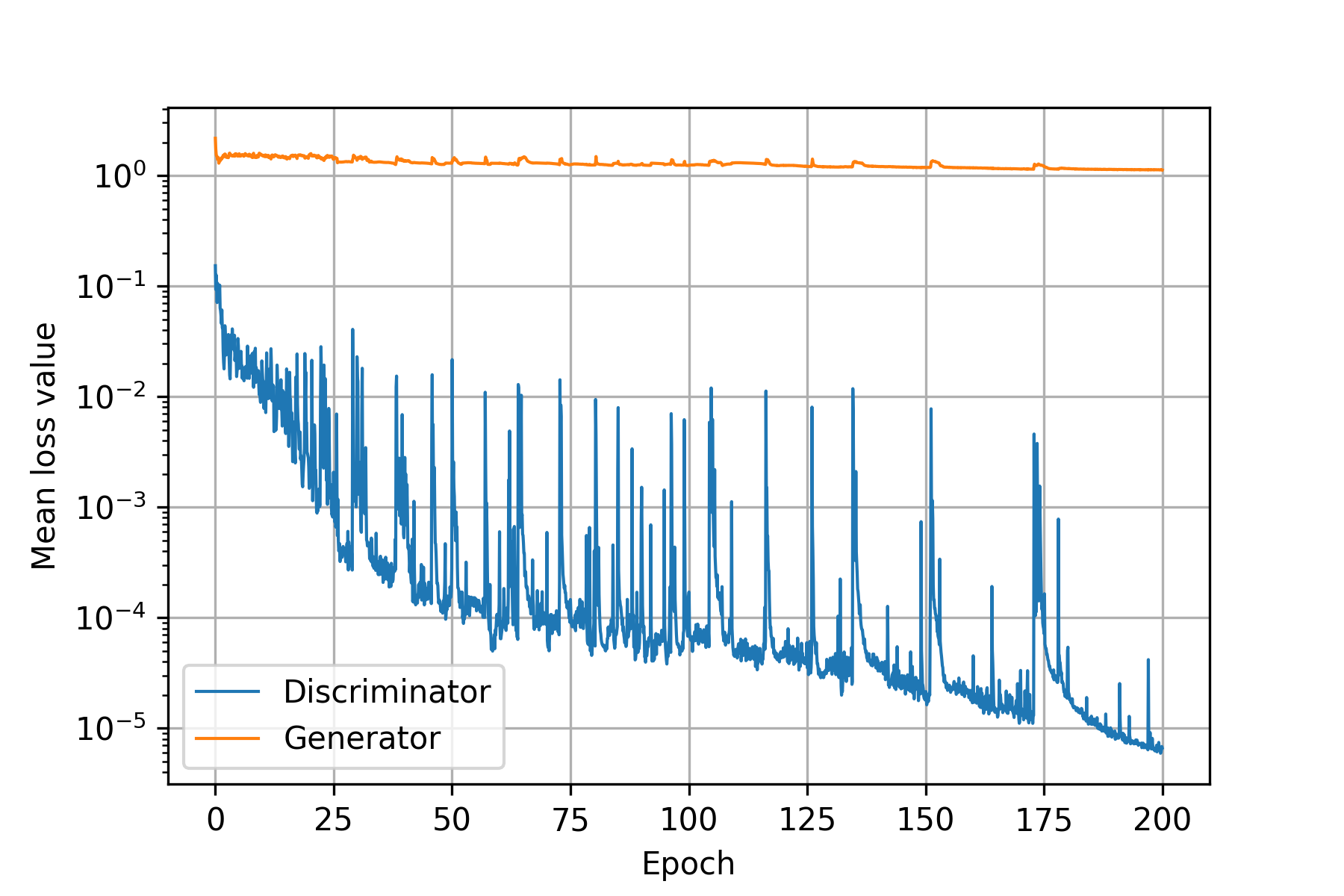}
    \caption{pix2pix}
    \label{fig9a:losses}
\end{subfigure}
\begin{subfigure}{0.3\linewidth}
    \centering
    \includegraphics[width=\linewidth]{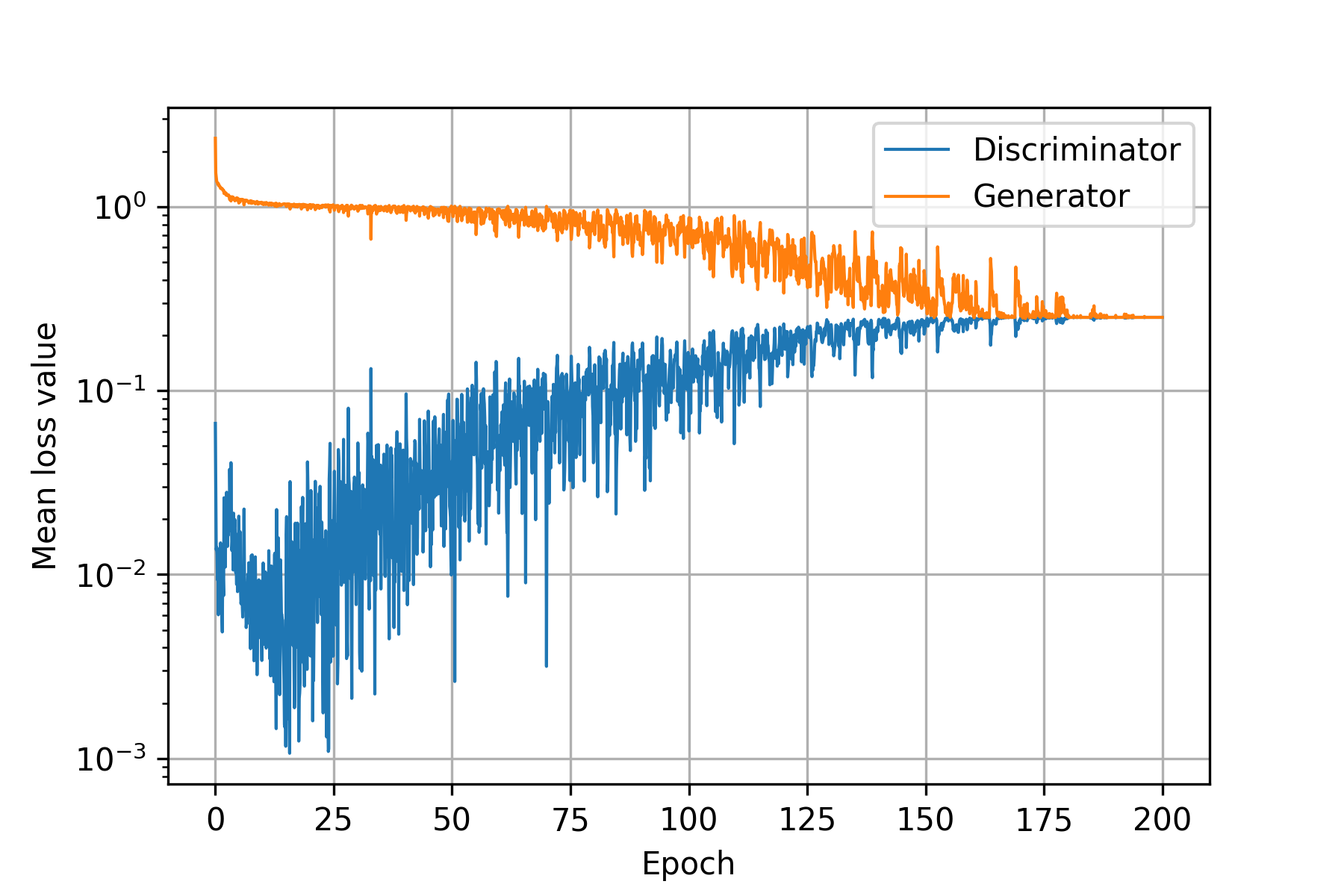}
    \caption{+ cross-entropy}
    \label{fig9b:losses}
\end{subfigure}
\begin{subfigure}{0.3\linewidth}
    \centering
    \includegraphics[width=\linewidth]{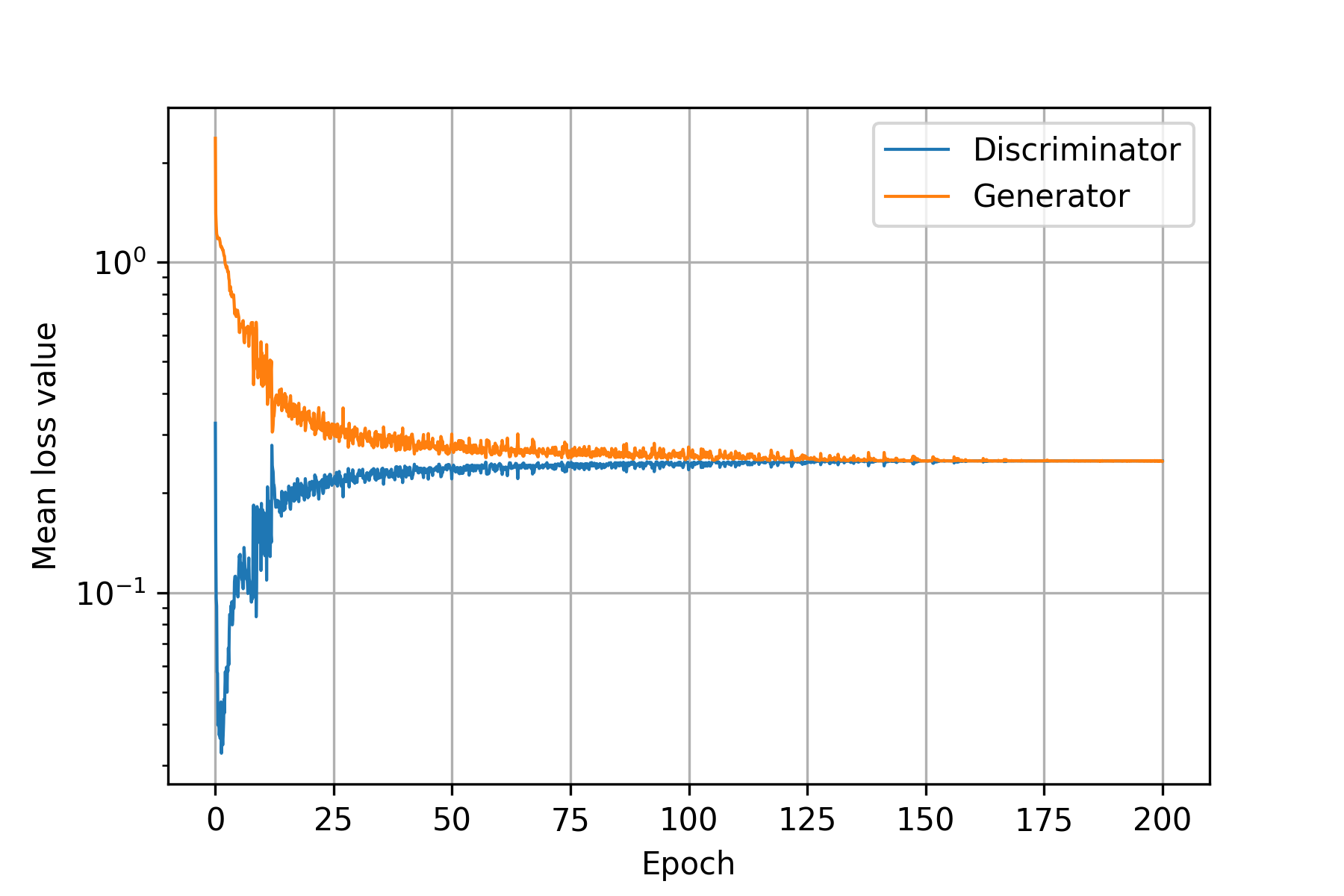}
    \caption{GAN-finder} 
    \label{fig9c:losses}
\end{subfigure}
\caption{Generator and Discriminator losses (in blue and orange respectively) on the grids with 20\% of rectangular obstacles.}
\label{fig9:losses}
\end{figure}

\subsubsection{Discriminator} 
The task of the discriminator is to detect whether the generated image comes from the distribution of the ground-truth images or not. We opt to focus discriminator only on the path component of the image, i.e. to train it to detect fake paths, thus we modify the input of the discriminator to be one-channel image which contains path pixels only. Such a ``simplification'' is evidently beneficial at the learning phase as otherwise the discriminator's loss is converging too fast and have no impact on the loss of the generator (see \figurename \ref{fig9:losses} on the left). Another reason for the discriminator to be focused on the path apart from obstacles is that the displacements of path pixels (e.g. putting them inside the obstacles) is penalized by the supervised part of the generator (i.e., via cross-entropy loss) thus the discriminator should rather detect how well the sequence of cells resembles the path pattern in general (are all cell adjacent, are there no loops/gaps etc.). Such an approach also naturally aligns with the idea that there may exist a handful of equivalent and plausible (and even optimal) paths from start to goal while the ground truth image depicts only one of them. 

In contrast to \cite{DBLP:journals/corr/IsolaZZE16} we achieved the best performance when training discriminator un-conditionally (without using input image as a condition). Implementing gradient penalty using Wasserstein distance \cite{DBLP:journals/corr/GulrajaniAADC17} for training the discriminator also yields better results. 

\subsubsection{Image post-processing}
To make sure that all obstacles remain at their places we transfer all blocked pixels from the original image to the generated one. Another technique we opt to use is gap-filling. We noticed that often a generated path is missing some segments. We use Bresenham line-drawing algorithm \cite{Bresenham:1965:ACC:1663347.1663349} to depict them. If the line segment is drawn across the obstacle, the gap remains.

\subsubsection{Success metrics}
In the image domain one of the most common metrics used to measure how well the model generates images is per-pixel mean-squared error (MSE). Path finding scenario is different in a sense that generated path pixels may be put to the other places compared to the ground truth image, thus MSE will be high, although the result is plausible, as it simply depicts some alternative path -- see \figurename \ref{fig_diff_paths}. We have already accounted for that fact when forced the discriminator to focus on the path structure in general rather than on the path pixels to be put precisely to the same places as on the ground truth image. We now want to account for it at test time so we introduce an additional metric called ``gaps'' which measures how many gaps are present in the generated path before post-processing. Finally, after the gaps are attempted to be removed we count the result as the ``success'' in case there are none of them left and the path truly connects start with goal (thus the path finding instance is solved).

\section{Experimental evaluation}
\subsubsection{Dataset}
We evaluated the proposed GAN on the $64 \times 64$ grids, representing outdoor environments with obstacles, as those types of grids were used in previous research on application of deep learning techniques to path finding (see \cite{tamar2016value}, \cite{lee2018gated} for example). Start was always close to the left border, goal -- to the right (as any grid can be rotated to fit this template). We used two approaches to put obstacles. In the first approach, rectangular obstacles of random size and orientation were put to random positions on the grid until obstacle density reached $20\%$ (we also used maps with $30\%$ for the evaluation but not learning). In the second approach, obstacles of rectangular, diamond and circular shape of random size were put to the map until the random obstacle density in the interval $[0.05; 0.5]$ was reached. The total size of each dataset was $50 000$, divided into train -- 75 \%, test -- 15 \% and validation -- 10 \%. For each input we built a ground-truth image depicting $A*$ path.

\begin{table}[t]
\centering
\caption{Success metrics on different types of data.}
\label{tab1}
\begin{tabular}{|c|ccc|ccc||ccc|}
\hline
&\multicolumn{3}{|c|}{20 \% density}&\multicolumn{3}{|c||}{30 \% density}&\multicolumn{3}{|c|}{Random}\\
\hline
&MSE & Gaps & Success &MSE & Gaps & Success &MSE & Gaps & Success\\
\hline
pix2pix \cite{DBLP:journals/corr/IsolaZZE16}& 0.0336 &19.54&65\%& 0.13& 27.22& 57\%& 0.2 & 27.56& 32\%\\
GAN-finder &0.014&1.4916&91.4\%&0.164 &2.71&73.1\%&0.045&3.142& 65.1\%\\
\hline
\end{tabular}
\end{table}

\begin{figure}[t]
\centering
\begin{subfigure}{\textwidth}
    \centering 
    \begin{subfigure}{0.49\textwidth}
        \centering
        \includegraphics[width=\linewidth]{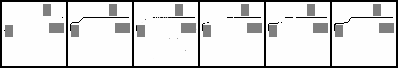}
        \caption{20 \% density}
        \label{fig10:20_den}
    \end{subfigure}
    \begin{subfigure}{0.49\textwidth}
        \centering
        \includegraphics[width=\linewidth]{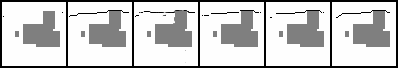}
        \caption {30 \% density}
        \label{fig10:30_den}
    \end{subfigure}
\end{subfigure}
\begin{subfigure}{\textwidth}
\centering
     \includegraphics[width=\linewidth]{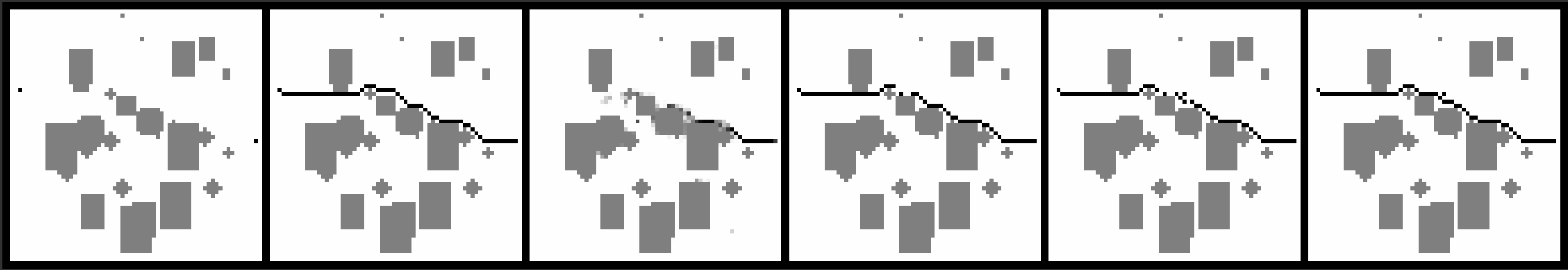}
    \caption{ Random data evaluation}
    \label{fig10:random}
\end{subfigure}
    \caption{From left to right: (1) input, (2) ground truth, (3) baseline pix2pix, (4) pix2pix using cross-entropy, (5) GAN-finder output, (6) GAN-finder output post-processed.}
\label{fig10:all}
\end{figure}

\subsubsection{Evaluation}

Figure \ref{fig9:losses} illustrates the training process for (a) baseline pix2pix GAN, (b) pix2pix trained with cross-entropy and (c) GAN-finder. It is clearly seen that GAN-finder converges much faster and in a more stable fashion. The examples of the paths found by various modifications of the considered GANs are shown in \figurename \ref{fig10:all}.

Success metrics for the 20 \% density maps (test part of the dataset) are shown in table \ref{tab1} on the left. We also evaluated the trained model on the 30 \% density maps (maps with such density were not part of the training) -- results are shown in table \ref{tab1} in the middle. Observing these results, one can claim that GAN-finder adapts well to the unseen instances with the same obstacle density (success rate exceeds 90\%). It is also capable to adapt to the unseen instances of higher obstacle density. Although in this case success rate is notably lower (around 73\%), it does not degrade to near zero values, which means that the model has indeed learned some general path finding techniques. One may also notice that GAN-finder significantly reduces the number of gaps (up to an order of magnitude), compared to baseline. The results achieved on the random dataset are shown in the right column of the table \ref{tab1}. Again GAN-finder is performing much better than the baseline. At the same time, success rate is now lower compared to 20\% density maps. We believe this is due to the more complex structure of the environments. One possible way to increase the success rate in this case might be to use more samples for training, another -- to use attention/recurrent blocks as in \cite{gregor2015draw}. Overall, the results of the evaluation are convincing enough to claim that the suggested approach, i.e. using GANs for path planning, is worth further investigation.

\section{Conclusion and Future Work}
In this work we suggested the generative adversarial network -- GAN-finder -- capable of solving path finding problems via image generation. Obtained results, being preliminary in nature, demonstrate that the suggested approach has a potential for further development as clearly the neural net has learned certain path finding basics. We are planning to extend this work in the following directions.

First, we want to study GAN-finder behaviour in more complex domains (e.g. the ones populated with complex-shaped obstacles, with highly varying obstacle densities etc.). As well we need to fairly compare our method with the other learning-based approaches such as Value Iteration Networks \cite{tamar2016value}.  

Second, we wish to further enhance the model to make it more versatile tool suitable for path planning. One of such enhancements is modifying the generator's loss in correlation with the idea of multiple possible paths from start to goal. E.g., we can ignore path pixels when computing cross-entropy loss but introduce an extra semi- or non-supervised loss component for them in addition  to (or completely substituting) the discriminator's feedback. Another appealing option is to add attention/recurrent blocks to the model. This will provide a capability to successively refine the path in complex domains, e.g. the ones densely populated with the obstacles of non-trivial shapes. This also might help in scaling to large maps.

Finally, we can use GAN-finder not just as a path planner on it's own, but rather as a complimentary tool for the conventional and well-established heuristic search algorithms, e.g. A*, providing them with a more informed heuristics.

\subsubsection{Acknowledgements} This work was supported by the Russian Science Foundation (Project No. 16-11-00048)

\bibliographystyle{splncs04}

\end{document}